\documentclass[10pt,twocolumn,letterpaper]{article}

\usepackage{iccv}
\usepackage{times}
\usepackage{epsfig}
\usepackage{graphicx}
\usepackage{amsmath}
\usepackage{amssymb}

\usepackage{subfigure} 
\usepackage{algorithm}
\usepackage{verbatim}
\usepackage{relsize}
\usepackage{algorithmicx}
\usepackage{algpseudocode}
\floatname{algorithm}{Algorithm}

\usepackage{amsthm}

\usepackage{subfigure}
\usepackage{bm}
\usepackage{multirow}
\newcommand{\vhat}[1]{{\bm{\hat #1}}}
\newcommand{\hatx}{\vhat{x}}
\newcommand{\haty}{\vhat{y}}
\newcommand{\hatz}{\vhat{z}}
\newcommand{\x}{\bm{x}}
\newcommand{\y}{\bm{y}}
\newcommand{\z}{\bm{z}}
\newcommand{\bo}{\bm{o}}

\newcommand{\lam}{\bm\lambda}

\usepackage[pagebackref=true,breaklinks=true,letterpaper=true,colorlinks,bookmarks=false]{hyperref}
\usepackage{amsfonts}

\iccvfinalcopy 

\ificcvfinal\fi

\begin{document}

    \title{HLIC: Harmonizing Optimization Metrics in Learned Image Compression by Reinforcement Learning}

    \author{Baocheng Sun \quad Meng Gu \quad Dailan He \quad Tongda Xu \quad Yan Wang\thanks{Corresponding author. This work is done when  Baocheng Sun and Meng Gu are interns at SenseTime Research.} \quad Hongwei Qin\\
        SenseTime Research\\
        {\tt\small \{sunbaocheng,  gumeng, hedailan, xutongda, wangyan1, qinhongwei\}@sensetime.com}
    }

    \maketitle
    \ificcvfinal\thispagestyle{empty}\fi

    \begin{abstract}
        Learned image compression is making good progress in recent years. Peak signal-to-noise ratio (PSNR) and multi-scale structural similarity (MS-SSIM) are the two most popular evaluation metrics. As different metrics only reflect certain aspects of human perception, works in this field normally optimize two models using PSNR and MS-SSIM as loss function separately, which is suboptimal and makes it difficult to select the model with best visual quality or overall performance. Towards solving this problem, we propose to Harmonize optimization metrics in Learned Image Compression (HLIC) using online loss function adaptation by reinforcement learning. By doing so, we are able to leverage the advantages of both PSNR and MS-SSIM, achieving better visual quality and higher VMAF score. To our knowledge, our work is the first to explore automatic loss function adaptation for harmonizing optimization metrics in low level vision tasks like learned image compression.
    \end{abstract}

    \section{Introduction}
    \label{sec:intro}
    With the growth of information technology, image capturing and sharing have become ubiquitous. 
    The purpose of image compression is to reduce storage and transmission cost.

    The basic idea of image compression is to reduce image signal redundancies by considering spatial correlations, statistical representations and human vision sensitivity. Traditional image compression methods, such as JPEG~\cite{ITU1992Information}, JPEG2000~\cite{rabbani2002jpeg2000}, BPG~\cite{bellard2015bpg} and the intra coding of VVC/VTM~\cite{vtm2019, ISOVVC}, all rely on hand-crafted modules. They incorporate intra prediction, discrete cosine transform/wavelet transform, quantization and entropy coder such as Huffman coder or content adaptive binary arithmetic coder (CABAC)~\cite{marpe2003CABAC} to remove such redundancies. 

    In recent years, deep learning and computer vision methods are introduced to create novel and powerful image compression techniques~\cite{rippel2017real, toderici2017full, balle2018variational, li2018learning, choi2019variable, cheng2020learned}. They utilize modern neural networks to obtain compact image representations, which can be subsequently quantized and compressed by standard entropy coding algorithms. Bit rate and image quality are constrained by loss function. Various entropy models~\cite{toderici2017full, balle2016end, mentzer2018conditional, balle2018variational, cheng2020learned} are designed as differentiable proxy to bit rate, proved to be very effective. The assessment of image quality is still an open problem. In image and video coding tasks, peak signal-to-noise ratio (PSNR) and multi-scale structural similarity (MS-SSIM)~\cite{wang2003multiscale} are two most common metrics. Both of them are differentiable and can be adopted as the distortion part in loss function. The tradeoff between bit rate and distortion can be controlled by adjusting the weight between entropy loss and distortion loss.
    
    It is known that fitting to different optimization metrics can cause different visual artifacts. And the choice of distortion loss is difficult~\cite{balle2018variational, NIPS2020_HIFIC}. 
    Works in the field of learned image compression tend to train two models optimizing PSNR and MS-SSIM separately. Models optimized for PSNR often have unsatisfactory performance on MS-SSIM, and vice versa~\cite{balle2018variational, minnen2018joint, cheng2020learned}. A linear hybrid of mean squared error (MSE, the only non-constant part in PSNR) and MS-SSIM has been used as perceptional loss in previous leading solutions to learned image compression challenge (CLIC)~\cite{zhou2018variational,  zhou2019end}. DSSIM proposed by~\cite{johnston2018improved} is designed as $L_1$ loss weighted by SSIM, which performs better than $L_1$ loss regarding PSNR, SSIM and MS-SSIM.
    Although these studies try to harmonize PSNR and MS-SSIM by hand-crafted loss function, deeper analysis and more automatic method are still lacking. For example, to what extent we can harmonize PSNR and MS-SSIM in one model, how to control the harmonization result conveniently, and how to do metric harmonization automatically or dynamically in the whole training process.
    
    In this paper, we aim to harmonize the optimization of PSNR and MS-SSIM in learned image compression by dynamically adapting the loss function using reinforcement learning (RL). Our contributions can be summarized as follows: 
    1) Our work is the first to investigate automatic loss function adaptation for metric harmonization in low level vision tasks like learned image compression.
    2) We propose an effective framework for controlling metric harmonization, which achieves improved and controllable tradeoff between PSNR and MS-SSIM. 
    3) We demonstrate the improvement brought by harmonious optimization on the famous perceptual metric VMAF~\cite{vmaf} and human perceptual quality, bringing new insight into image quality optimization for low level vision tasks.

    \section{Related Work}

    \subsection{Learned Image Compression}
    A number of powerful learned image compression methods have emerged with the rapid development of deep learning~\cite{balle2020nonlinear}. Ball\'{e} \etal~\cite{balle2016end} propose an autoencoder based structure to perform nonlinear transform coding of images. The output of encoder is quantized and viewed as latent code to be compressed by lossless entropy coders. 
    Later,~\cite{balle2018variational} extends this model by utilizing hyperprior to capture the interdependency of latent representation, leading to a much more powerful entropy model. Inspired by PixelCNN~\cite{pixelcnn}, the hyperprior model is extended by adding an autoregressive module to further exploit the probabilistic structure of the latent code~\cite{minnen2018joint}. 
    Cheng \etal~\cite{cheng2020learned} propose to use more powerful neural network architectures and use Gaussian Mixture Model (GMM) for entropy modelling, which is the first work comparable with the intra coding of Versatile Video Coding (VVC)~\cite{ISOVVC} regarding PSNR. 
    
    As stated in the introduction, works in this field normally train two separate models optimizing PSNR and MS-SSIM respectively. A few studies investigate the harmonization of different metrics by hand-crafted loss function~\cite{zhou2018variational, zhou2019end, johnston2018improved}. However, as mentioned above, there are still problems to be investigated.

    \subsection{Visual Quality Optimization}

    PSNR is the de facto standard for traditional codec comparison~\cite{vmaf}. However, this pixel-based difference measurement correlates poorly with human perception~\cite{girod1993s, DLM}.
    MS-SSIM is another popular and differentiable metric widely used in image compression. While not directly related to human perception, it obtains better subjective ratings than PSNR. Images obtained by optimizing PSNR have clearer structural information, and images obtained by optimizing MS-SSIM retain more texture~\cite{balle2018variational, LeeCGAN}. 

    In recent years, VMAF~\cite{vmaf} uses Support Vector Machine to fuse a number of elementary metrics, such as VIF~\cite{VIF} and DLM~\cite{DLM}. It achieves good subjective rating results and has become an industry standard for video coding. Unfortunately, It is not differentiable and cannot be used directly for optimizing learned image compression. ProxIQA~\cite{chen2020proxiqa} proposes a proxy network to mimic VMAF, which is used as loss function for learned image compression, leading to 20\% bitrate reduction to MSE-optimized baseline under the same VMAF value.
    
    VGG based loss has been used in diverse tasks as perceptual constraint~\cite{johnson2016perceptual, ledig2017photo}. 
    However, it does not perform well in learned image compression due to unpleasant artifacts~\cite{NIPS2020_HIFIC}.

    \subsection{Loss Function Search/Adaptation}
    Loss function search is an AutoML technology and becomes popular recently. 
    By dynamically optimizing the parameters of loss function's distribution using REINFORCE algorithm~\cite{williams1992simple}, AM-LFS~\cite{amlfs} surpasses hand-crafted loss functions in classification, face recognition and person ReID. Wang \etal~\cite{ShifengFace} further improves this method for face recognition by redesigning the search space based on deeper analysis of margin-based softmax losses. 
    Very recently, Ada-Segment~\cite{zhang2020ada} extends the idea of AM-LFS to perform multi-loss adaptation for panoptic segmentation. 
    Auto Seg-Loss~\cite{li2020auto} proposes to search for parameterized metric-specific surrogate loss in semantic segmentation. 
    CSE-Autoloss~\cite{liu2021loss} proposes to discover novel loss functions for object detection automatically by searching combinations of primitive mathematical operations with evolutionary algorithm.
    
    Existing works mainly focus on high-level vision tasks including classification, object detection, face recognition, person ReID and segmentation. However, those high-level tasks have a large difference with image compression, a typical low-level task. In high-level tasks, mis-alignment between loss function and evaluation metrics degrades the performance, leaving relatively larger room for improvement by loss function search or adaptation methods.
    In image compression, popular evaluation metrics including PSNR and MS-SSIM can be used as loss function. So it is much more difficulty to achieve performance improvement regrading these two metrics. In addition, in low level vision tasks no single evaluation metric or known combinations can well represent the ultimate goal of optimizing human visual perception, which brings additional complexity.
    

    \section{Proposed Method}



    \subsection{Differentiable Rate-Distortion Optimization}

    \begin{figure}
        \centering
        \includegraphics[width=8.5cm]{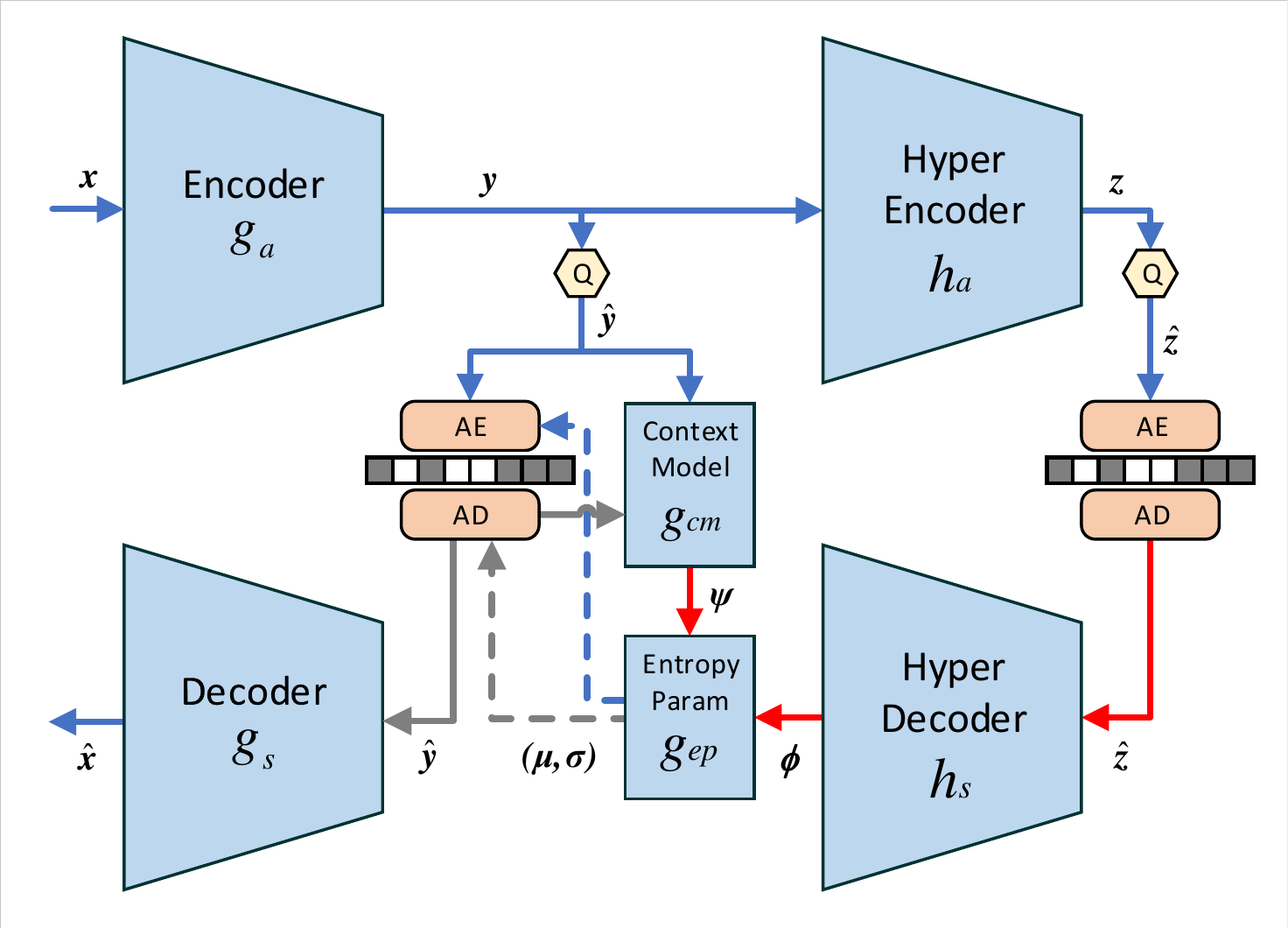}
        \caption{Diagram of autoregressive compression models~\cite{minnen2018joint, cheng2020learned}. The blue and gray arrows denote encoding and decoding data flow respectively while the red ones are shared by both encoding and decoding process.}
        \label{fig:arch-gg18}
    \end{figure}
    
    We formulate image compression as a differentiable rate-distrotion (RD) optimization problem following previous works~\cite{balle2018variational, minnen2018joint, cheng2020learned}. Figure~\ref{fig:arch-gg18} shows a typical model architecture. 
    First, the input image $\x$ is fed into a neural encoder $g_{a}$ and transformed into latent representation $\y$. Then we quantize it to get the code $\haty$ to be compressed. Since the rounding operator has zero gradients everywhere, additive uniform noise $\bm{q} \sim \mathcal{U}\left(-\frac{1}{2}, \frac{1}{2}\right)$ is introduced as a relaxation, generating $\tilde{\y} = \y + \bm{q}$ which approximates $\haty$ during training~\cite{balle2016end, balle2018variational}. 
    For simplicity, below we use $\haty$ to denote both $\haty$ and $\tilde{\y}$.
    Side-information $\z$ is further extracted from $\y$ by a hyper encoder $h_{a}$. Then it is quantized as $\hatz$ and get compressed into bitstream accompanied with $\haty$. The same relaxation is applied to generate $\tilde{\z}$ in training.
    During decoding, side-information $\hatz$ is decompressed from the bitstream first and transformed by the hyper decoder $h_s$ to predict the distribution $p_{\haty | \hatz}$. After that, $\haty$ can be decompressed and finally the output image $\hatx$ can be reconstructed from $\haty$ by transform $g_s$.
    
    The distribution $p_{\haty | \hatz}$ can be modelled as:
    \begin{equation}
        \begin{aligned} 
            p_{\haty|\hatz}(\haty) &= \prod_i \left( \mathcal{N}(\mu_i, \sigma^2_i) * \mathcal{U}(-\frac{1}{2}, \frac{1}{2}) (\hat{y}_i) \right)
        \end{aligned}
    \end{equation}
    where $i$ indicates the element id in $\haty$ and $\mu_i$ is set to $0$ in~\cite{balle2018variational}. The probability of $\vhat z$ is modeled using a non-parametric fully factorized density model proposed in~\cite{balle2016end}. The probability model for $\haty$ and $\hatz$ is used in the compression and decompression process of entropy coding. Also, they are part of the loss function to enable differentiable RD tradeoff, where the bit rate is approximated using entropy:
    \begin{equation}
        \begin{aligned}
            L &= R + D \\ 
            & = \mathbb{E}_{\x \sim p_{\x}}[-\log_2 p_{\vhat y|\vhat z}(\vhat y | \vhat z) -\log_2 p_{\vhat z}(\vhat z) ] \\
            &+ \lambda \cdot \mathbb{E}_{\bm x \sim p_{\bm x}} [d(\bm x, \vhat x)]
        \end{aligned}
        \label{eq:r-d}
    \end{equation} 
    where distortion weight $\lambda$ controls the RD trade-off. As introduced in section~\ref{sec:intro}, distortion term $d(\x, \hatx)$ is usually $\text{MSE}$ or $1-\text{MS-SSIM}$.
    
    Minnen \etal ~\cite{minnen2018joint} propose an autoregressive method to better estimate $p_{\haty | \hatz}$, which can be formulated as:
    \begin{equation}
        \begin{aligned}
            \bm\psi_i &= g_{cm}(\haty_{<i}) \\
            \bm\phi &= h_s(\hatz) \\
            \mu_i, \sigma_i &= g_{ep}(\bm\phi, \bm\psi_i) \\
            p_{\haty|\hatz} &= \prod_{i}\left( \mathcal{N}(\mu_i, \sigma^2_i) * \mathcal{U}(-\frac{1}{2}, \frac{1}{2}) \right) (\hat{y}_i)
        \end{aligned}
    \end{equation}
    where $\vhat y_{<i}$ means the context of $\hat y_i$ (\ie some accessible neighbours), $g_{cm}$ and $g_{ep}$ are also transforms using neural networks.
    Cheng \etal ~\cite{cheng2020learned} extends this method by using Gaussian Mixture Model to estimate $p_{\haty | \hatz}$:
    \begin{equation}
        \begin{aligned}  
            p_{\haty | \hatz} = \prod_{i} \left\{\sum_{0<k<K} \pi_i^{(k)} \left[ \mathcal{N}(\mu^{(k)}_i, \sigma^{2(k)}_i) * \mathcal{U}(-\frac{1}{2}, \frac{1}{2})\right](\hat{y}_i) \right\}
        \end{aligned}
    \end{equation}
    where K groups of entropy parameters $(\bm \pi^{(k)}, \bm \mu^{(k)}, \bm \sigma^{(k)})$ are generated by $g_{ep}$.




    \subsection{Metric Harmonization by Loss Function Adaptation}
    \subsubsection{Adaptation Space}

    Based on analysis above, the adaptation space of distortion loss $D$ can be formulated as:
    \begin{equation}
        \begin{aligned}
            D = \lambda_{\text{MSE}} \cdot \text{MSE} + \lambda_{\text{MS-SSIM}} \cdot (1 - \text{MS-SSIM})
        \end{aligned}
        \label{eq:d_space}
    \end{equation}
    where ${\text{MSE}}$ and $1 - {\text{MS-SSIM}}$ are the loss function for optmization metric PSNR and MS-SSIM respectively.
    As $\lambda_{\text{MSE}}$ and $\lambda_{\text{MS-SSIM}}$ are always positive, we use the trick of reparameterization:
    \begin{equation}
        \begin{aligned}
            \lambda_{\text{MSE}} &= e ^ {\lambda'_{\text{MSE}}} \\
            \lambda_{\text{MS-SSIM}} &= e ^ {\lambda'_{\text{MS-SSIM}}} \\
        \end{aligned}
        \label{eq:exp_lambda}
    \end{equation}
    where $ \lambda'_{\text{MSE}}$ and $\lambda'_{\text{MS-SSIM}}$ are the parameters to be adapted online (as a sequence of values) throughout the training process.
    We observe that in eq.~\ref{eq:r-d} the change in distortion weight $\lambda$ and the final bit rate forms a logarithm relationship. So in eq.~\ref{eq:exp_lambda} changes in $\lambda'_{\text{MSE}}$ and $\lambda'_{\text{MS-SSIM}}$ are almost proportional to changes in bit rate. Such mapping makes it easier for RL to converge on desirable $\lambda'_{\text{MSE}}$ and $\lambda'_{\text{MS-SSIM}}$ corresponding to different target bit rates. 

    \subsubsection{Reward Design}
    \begin{figure}
    \includegraphics[width=8.5cm]{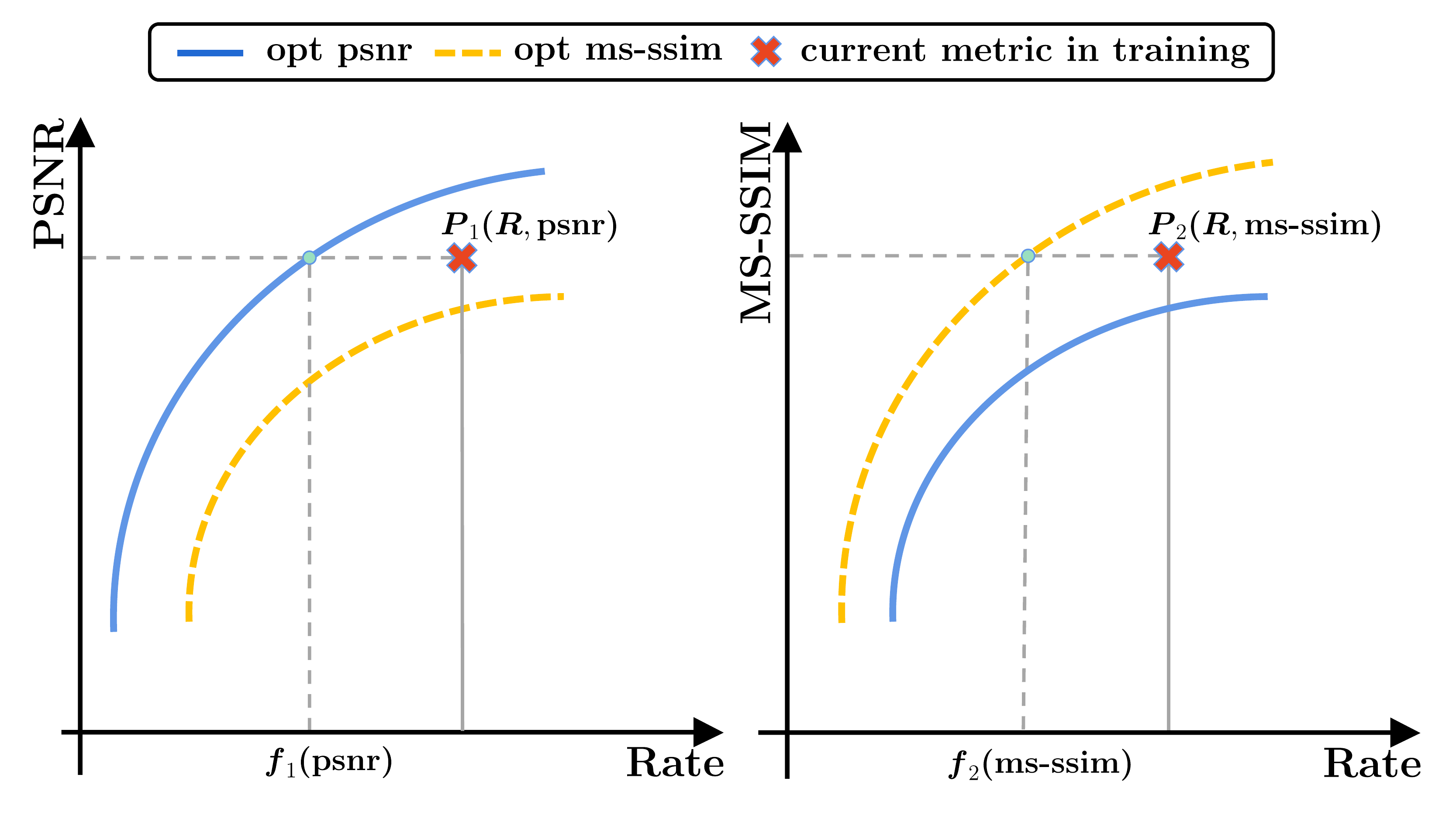}
    \caption{
         Mapping functions in the reward design. Blue and yellow RD curves denote baselines trained using MSE and MS-SSIM as distortion loss respectively. Red mark means current $(R, \text{psnr})$ and $(R, \text{ms-ssim})$.
    }
    \label{fig:reward}
\end{figure}
    Reward design is the key of controllable metric harmonization and improvement. Our reward consists of three parts and depends on two baseline RD curves optimized for PSNR and MS-SSIM separately. 
    Denote the model parameterized by $\theta$ as $\mathcal{M}_{\theta}$ and the validation set as $\mathcal{S}_v$. The reward can be formulated as a function of $\mathcal{M}_{\theta}$ and $\mathcal{S}_v$:
    \begin{equation}
        \begin{aligned}
            \mathcal{R} &= r(\mathcal{M}_{\theta}; \mathcal{S}_v) \\
            &= \mathop{\mathcal{W}_1}(\frac{R - R_{target}}{R_{target}}) \\
            &+ \mathop{\mathcal{W}_2}(\frac{R - f_1(\text{psnr})}{f_1(\text{psnr})}) \\
            &+ \mathop{\mathcal{W}_3}(\frac{R - f_2(\text{ms-ssim})}{f_2(\text{ms-ssim})}) 
        \end{aligned}
        \label{eq:reward}
    \end{equation}
    where $R$ is current bits per pixel (bpp). The function $f_1$ and $f_2$ represent the mapping from current metric value (denoted as lower-case psnr and ms-ssim in eq.~\ref{eq:reward} and Figure~\ref{fig:reward}) to corresponding bpp value in the baseline RD curves. 
    The first part encourages the bit rate to be the target one, the second and the third parts encourage the model to converge to the best RD tradeoff regarding PSNR and MS-SSIM respectively. 
    For different purposes, the three targets can be weighted and reshaped using different functions $\mathop{\mathcal{W}_i}$. 
    
    We design two types of experiments to demonstrate that our method can control and improve metric harmonization. For the First type, $\mathop{\mathcal{W}_i}$ can be expressed as:
    \begin{equation}
        \begin{aligned}
        \mathcal{W}_1(x) &= w_1 x^2 \\
        \mathcal{W}_2(x) &= 0 \\
        \mathcal{W}_3(x) &= w_3 x
        \end{aligned}
        \label{eq:reward1}
    \end{equation}
    We set $w_1 = 25, w_3 = 10$, which encourages the model to focus on optimizing MS-SSIM (section~\ref{section: ablation}).
    For the second type,  $\mathop{\mathcal{W}_i}$ can be expressed as:
    \begin{equation}
        \begin{aligned}
           \mathcal{W}_i(x) = w_i x^2 \quad (i=1, 2 ,3) 
        \end{aligned}
        \label{eq:reward2}
    \end{equation}
    where $w_i$ is used to control the priority between PSNR and MS-SSIM (section~\ref{section: controllable}). To optimize PSNR while giving attention to MS-SSIM, we set $w_1 = 25, w_2 = 100, w_3 = 1$. To optimize MS-SSIM while giving attention to PSNR, we set $w_1 = 25, w_2 = 1, w_3 = 100$.
    
    \subsubsection{Optimization}
    \begin{algorithm}[!h]
        \caption{PPO-based Loss Function Adaptation}
        \label{algo}
        \begin{algorithmic}
            \Require 
            Initialized model $\mathcal{M}_{\theta^*_0}$, epoch $T$,  
            the number of samples $B$
            , trajectory length $N$
            \Ensure optimal model $\mathcal{M}_{\theta^*_{T}}$
            \State $r^i_0, \bo^i_1$ $\gets$ \Call{eval}{$\mathcal{M}_{\theta^*_0}$;$\mathcal{S}_v$}, $i\in \{1,...B\}$
            \For{$t$ $\gets$ $1$ to $T$}
            \For{$i$ $\gets$ $1$ to $B$} \Comment{train $B$ models separately}
                \State Sample distortion weights $\lam_t^i \sim \pi_{\omega_p}(\cdot|\bo_t^i)$
                \State $\mathcal{M}_{\theta_{t}^i} \gets \Call{train}{\mathcal{M}_{\theta_{t-1}^i}; \mathcal{S}_t, \bm\lambda_i}$
                \State $r_t^i, \bo_{t+1}^i$ $\gets$ \Call{eval}{$\mathcal{M}_{\theta_t^i}$;$\mathcal{S}_v$}
            \EndFor
            \If{$t$ Mod $N$ = $0$}
            \State Update $\omega_p \gets \text{argmax}_{\omega_p} \: \mathbb{E}_{\pi_{\omega_p}}[\sum_{i,t} r_t^i]$
            \State $\text{index} \gets \text{argmax}_{j \in [1, B]} r_t^j $
            \State $\mathcal{M}_{\theta_{t}^*}$ $\gets$ $\mathcal{M}_{\theta_{t}^{\text{index}}}$ 
            \EndIf
            \EndFor
            \State \Return $\mathcal{M}_{\theta^*_{T}}$
        \end{algorithmic}
    \end{algorithm}

    We denote the training set as $\mathcal{S}_t$, the validation set as $\mathcal{S}_v$, and the image as $\x$.
    Here we are adapting hyper-parameters $\lam = ( \lambda'_{\text{MSE}}, \lambda'_{\text{MS-SSIM}}) $ defined in eq.~\ref{eq:exp_lambda}  online for a specific model $\mathcal{M}_{\theta}$.
    Before the $t^{th}$ epoch of training, an observation $\bm o_t$ is made. 
    $\bm o_t$ contains validation results from $(t-1)^{th}$ epoch including MS-SSIM, PSNR, bpp of $\haty$, bpp of $\hatz$, gradient loss and total variation~\cite{mahendran2015understanding}. The online adapting policy $\pi_{\omega_p}(\lam|\bm o_t)$ is a distribution over $\lam$ parameterized by $\omega_p$. Our target of loss function adaptation is to maximize the expectation of reward $r(\mathcal{M}_{\theta}; \mathcal{S}_v)$ in eq.~\ref{eq:reward} and the weight $\theta$
    is obtained by minimizing the online adapted RD loss modified from eq.~\ref{eq:r-d} and eq.~\ref{eq:d_space}:

    \begin{equation}
        \begin{aligned}
         &\pi^* = \underset{\{\pi (\cdot | \bm o)\}}{\textbf{argmax}} \:
         \mathlarger{\mathbb{E}}_{\lam \sim \pi(\cdot | \bm o)}
         \Big[
         \underset{t}{\sum}
         r_t(\mathcal{M}_{\bm\theta^*(\lam)}; \mathcal{S}_v)
         \Big] \\
         \textbf{s.t.\quad} &\\
           &\begin{aligned}
            \bm\theta^*(\lam) = 
            \mathop{\textbf{argmin}}\limits_{\{
            \bm\theta
            \}} 
            \frac{1}{\#\mathcal{S}_t}\sum_{\x \in \mathcal{S}_t} (R 
            +\lambda_{\text{MSE}} \cdot \text{MSE}\\
             +\lambda_{\text{MS-SSIM}} \cdot (1- \text{MS-SSIM})) 
            \end{aligned}
        \end{aligned}
        \label{opt-target}
    \end{equation}
    where $\#\mathcal{S}_t$ denotes the size of training set.
    To solve this bi-level optimization problem, ~\cite{amlfs} adopts REINFORCE~\cite{williams1992simple}. However, REINFORCE suffers from unstable training due to high variance. 
    
    We tackle this problem with Proximal Policy Optimization (PPO)~\cite{schulman2017proximal}. 
    After observation $\bm o_t$ is made, we sample distortion weights $\lam_t$ 
    from policy $\pi_{\omega_p}(\cdot|\bm o_t)$ for $t^{th}$ epoch.
    Our policy is normal distribution with $\bm\mu_{\lam}$, $\bm\sigma_{\lam}$ controlled by a multi-layer perceptron (MLP):
    \begin{equation}
        \begin{aligned}
            \lam_t &\sim \pi_{\omega_p}(\cdot |\bo_t) \\
            \pi_{\omega_p}(\cdot|\bm o_t) &= \mathcal{N}(\bm\mu_{\lam}, {\bm\sigma^2_{\lam}} )\\
            & \text{with } \bm\mu_{\lam}, \bm\sigma_{\lam} = \mathcal{MLP}(\bm o_t)
            \label{sample}
        \end{aligned}
    \end{equation}
    After training $B$ models with distortion weight $\lam_t$ for $t^{th}$ epoch, we validate these models to obtain reward 
    $r_t = r(\mathcal{M}_{\bm\theta_t}; \mathcal{S}_v)$ 
    and the next observation $\bm o_{t+1}$. To reduce the variance of the gradient estimation, we utilize critic~\cite{KondaT99} and generalized advantage estimation~\cite{Schulmanetal_ICLR2016} to compute advantage $\hat{\bm A}_t$. 

    Every $N$ epoch of training, we generate $B$ trajectories of length $N$. Then, we optimize the parameters $\omega_p$ of policy $\pi_{\omega_p}$ by minimizing surrogate loss $L^{\textbf{CLIP}}$:
    \begin{equation}
        \begin{aligned}
            L^{\textbf{CLIP}}(\omega_p)& = \hat{\mathbb{E}}_t \left[ \text{min}\Big(
                f_t(\omega_p)\hat{\bm A}_t, \right. \\
                & \quad \qquad \left. \text{CLIP}\big(f_t(\omega_p), 1-\epsilon, 1+\epsilon\big)\hat{\bm A}_t
                \Big) \right] \\
            &\text{with } f_t(\omega_p)
            = \frac{\pi_{\omega_p}(\lam_t|\bm o_t)}{\pi_{\omega_{p_t}}(\lam_t|\bm o_t)}
        \end{aligned}
    \end{equation}
    where the clipping ratio $\epsilon$ is $0.2$ and $f_t$ is the important sampling factor. Furthermore, we broadcast the model $\mathcal{M}_{\bm\theta_{t}^*}$ with the highest reward $r(\mathcal{M}_{\bm\theta_t^*}; \mathcal{S}_v)$ to initialize the $B$ models in the next epoch.
    
    The full pipeline is presented in Algorithm \ref{algo}.

\section{Implementation Details}
\textbf{Training Details:}  
We investigate our proposed metric harmonization framework based on two representative learned image compression methods: Ball\'e2018~\cite{balle2018variational} and Cheng2020~\cite{cheng2020learned}. 
We use same network architectures following original paper.
We selected 8000 largest pictures from the Imagenet dataset~\cite{deng2009imagenet} as the training set $\mathcal{S}_t$. According to previous works~\cite{balle2016end, balle2018variational}, we add uniform random noise to each picture, and then perform down-sampling. During the training process, the picture is cropped into samples with the size of $256 \times 256$. All models are trained for 2000 epochs (\ie 1M steps) with a batch-size of 16 and use a constant learning rate of $1 \times 10^{-4}$ if not specified.
To optimize the policy, we use the Adam optimizer with a learning rate of $\eta_{\theta_p} = 0.0003$, $\beta_1 = 0.9 $ and $\beta_2 = 0.999$. The number of samples $B$ is set to 8. 
More detailed description is given in the appendix.

\textbf{Evaluation:} We use Kodak dataset~\cite{kodak} as our validation set $\mathcal{S}_v$ and Tecnick dataset~\cite{tecnick2014TESTIMAGES} as our test set. 

\textbf{Traditional Codecs:} We also compare our methods with VTM, BPG and JPEG. 

\begin{figure*}
    \begin{minipage}[t]{0.5\linewidth}
        \centering
        \includegraphics[width=8.5cm]{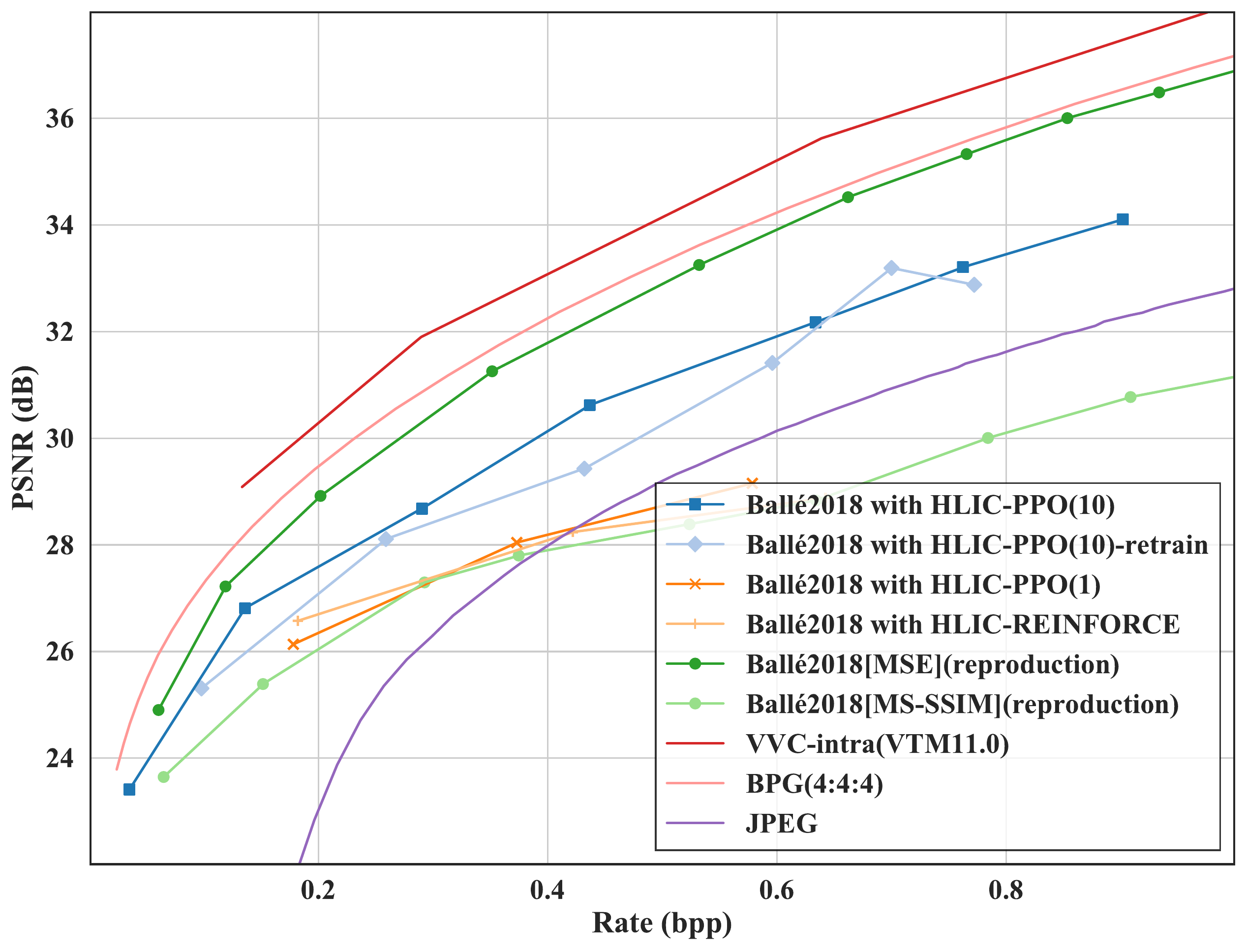}
    \end{minipage}
    \begin{minipage}[t]{0.5\linewidth}
        \centering
        \includegraphics[width=8.5cm]{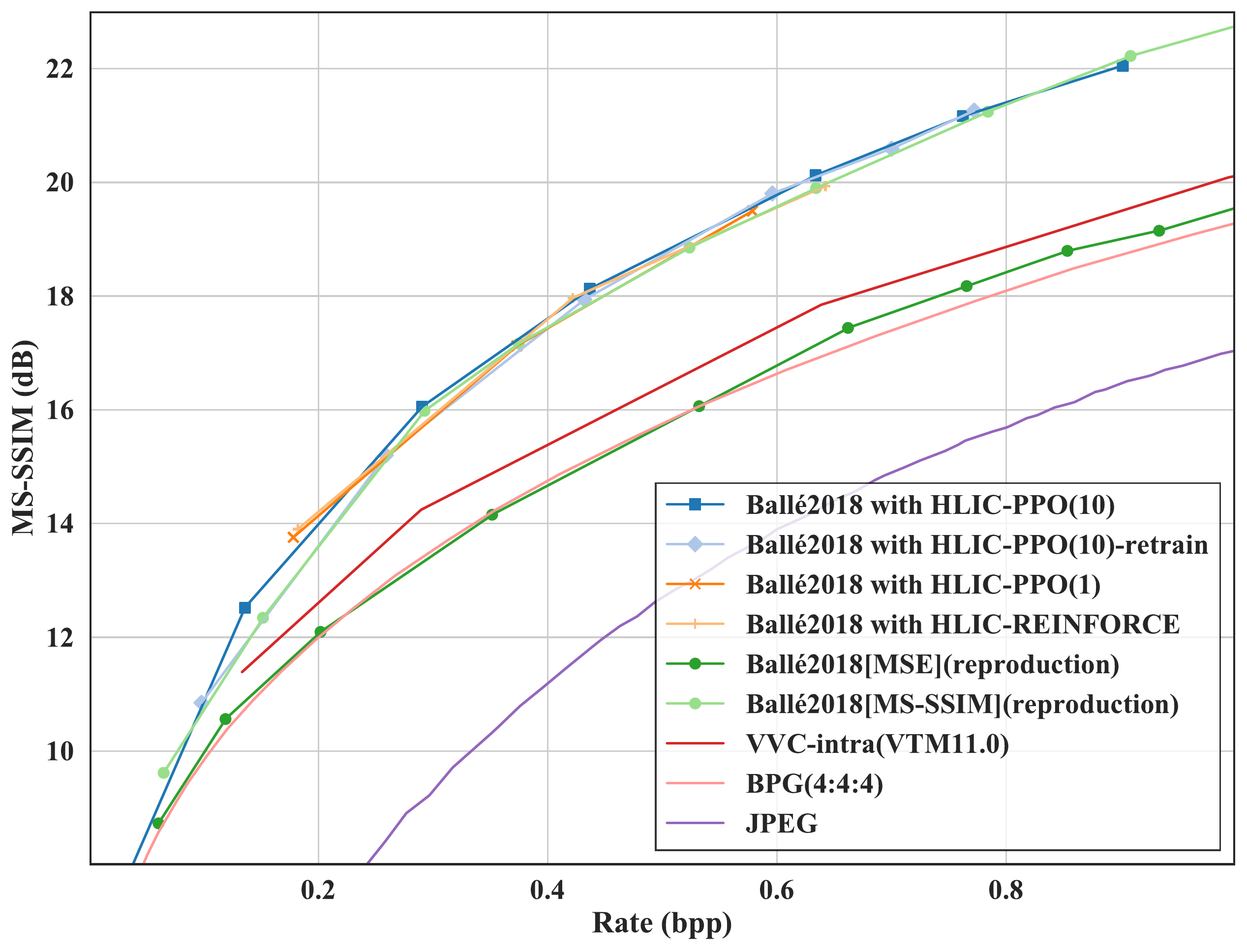}
    \end{minipage}
    \caption{
         Ablation Study On Kodak Dataset.
    }
    \label{fig:ablation_kodak}
\end{figure*}

\begin{figure*}
    \begin{minipage}[t]{0.5\linewidth}
        \centering
        \includegraphics[width=8.5cm]{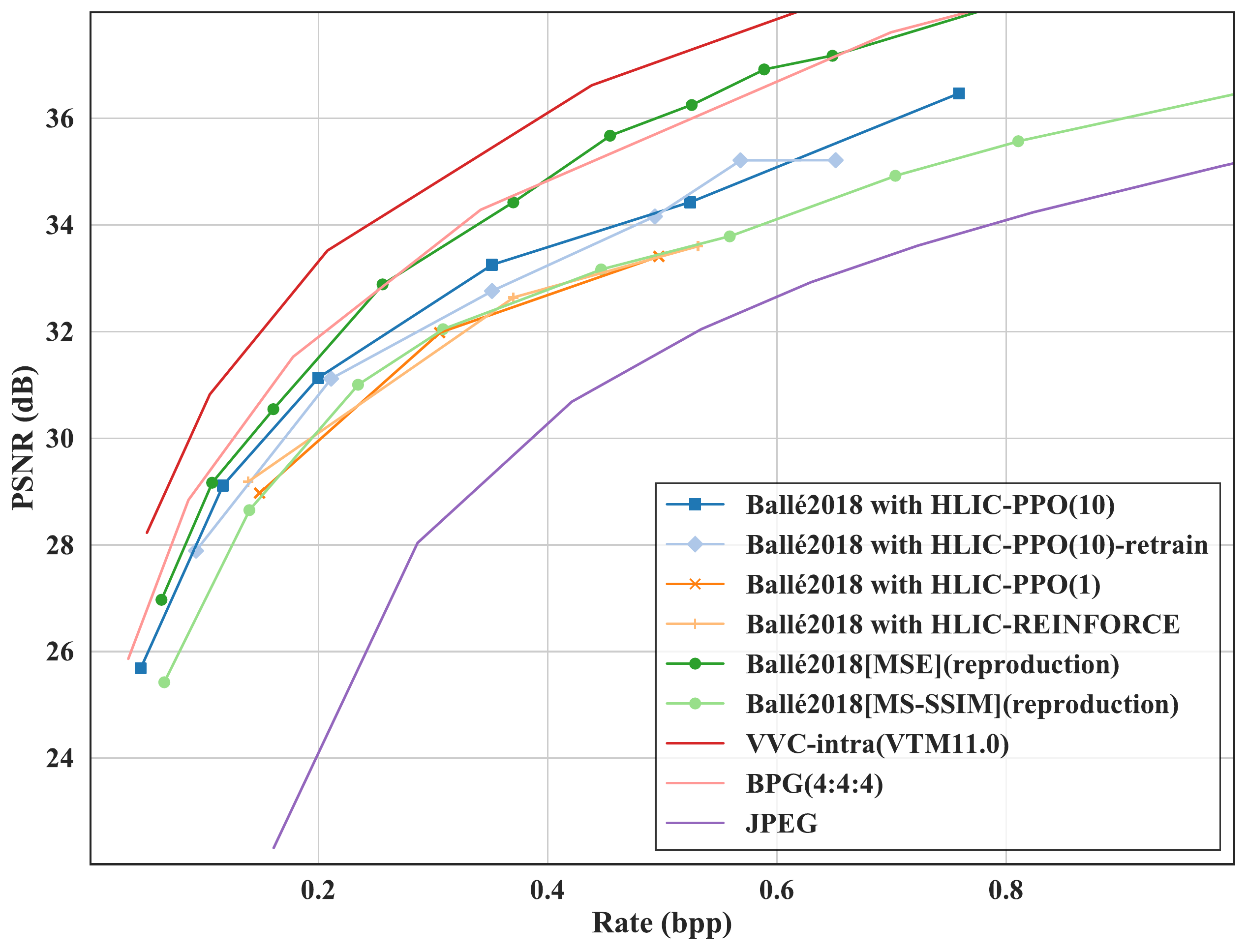}
    \end{minipage}
    \begin{minipage}[t]{0.5\linewidth}
        \centering
        \includegraphics[width=8.5cm]{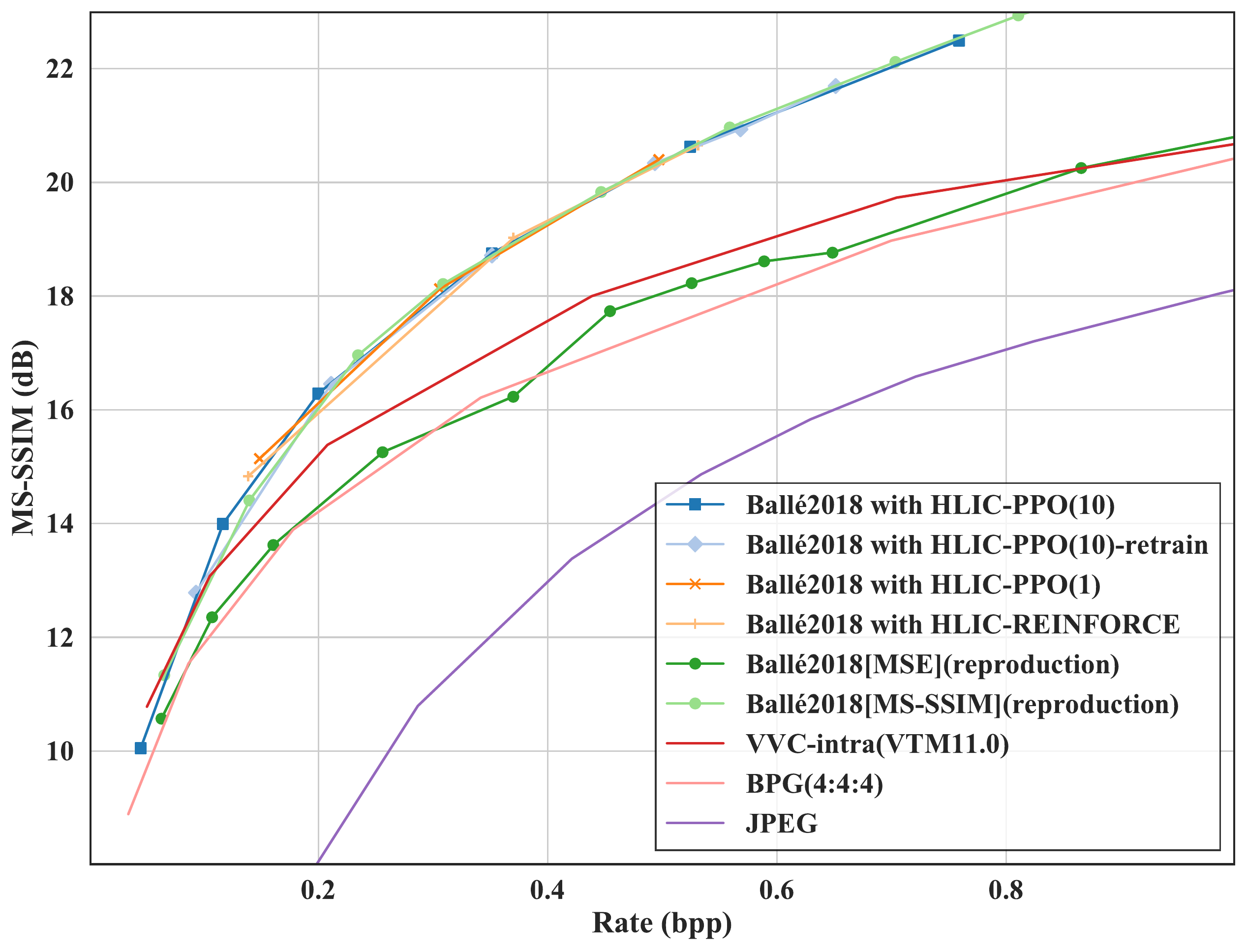}
    \end{minipage}
    \caption{
         Ablation Study On Tecnick Dataset.
    }
    \label{fig:ablation_tecnick}
\end{figure*}

\section{Experiments}
\subsection{Effectiveness and Ablation Study}
\label{section: ablation}
In order to show the effectiveness of our purposed method, we compare it with the most common training method for learned image compression, where only MS-SSIM is used as loss function (marked as Ball\'e2018[MS-SSIM](reproduction) in Figure~\ref{fig:ablation_kodak} and Figure~\ref{fig:ablation_tecnick}). 
We use the reward defined in eq.~\ref{eq:reward1}, where we only encourage MS-SSIM. 
We find that the RD performance not only improves on PSNR, but also slightly improves on MS-SSIM. Using the same reward, we also test PPO with a trajectory length of $1$ and REINFORCE for comparisons, but they are unable to achieve the result of PPO with a trajectory length of $10$. Previous works~\cite{balle2018variational, minnen2018joint, cheng2020learned} show that if only one metric is optimized, the test result of the other one will perform very poor. 
Our experimental results show that if MSE is utilized skillfully in the loss function, MS-SSIM can be further improved slightly, and at the same time PSNR can be improved apparently compared with only optimizing MS-SSIM baseline. This indicates that the optimization of PSNR and MS-SSIM can be harmonized to some extent. We take the adapted loss function at the last iteration of HLIC and use it as a fixed loss function to retrain a new model from scratch (marked with retrain in Figure~\ref{fig:ablation_kodak} and Figure~\ref{fig:ablation_tecnick}), which performs apparently worse than our HLIC method, indicating that the proposed online loss adaptation process benefits the harmonization of PSNR and MS-SSIM.

\subsection{Controllable Metric Harmonization}
\begin{figure*}
    \begin{minipage}[t]{0.5\linewidth}
        \centering
        \includegraphics[width=8.5cm]{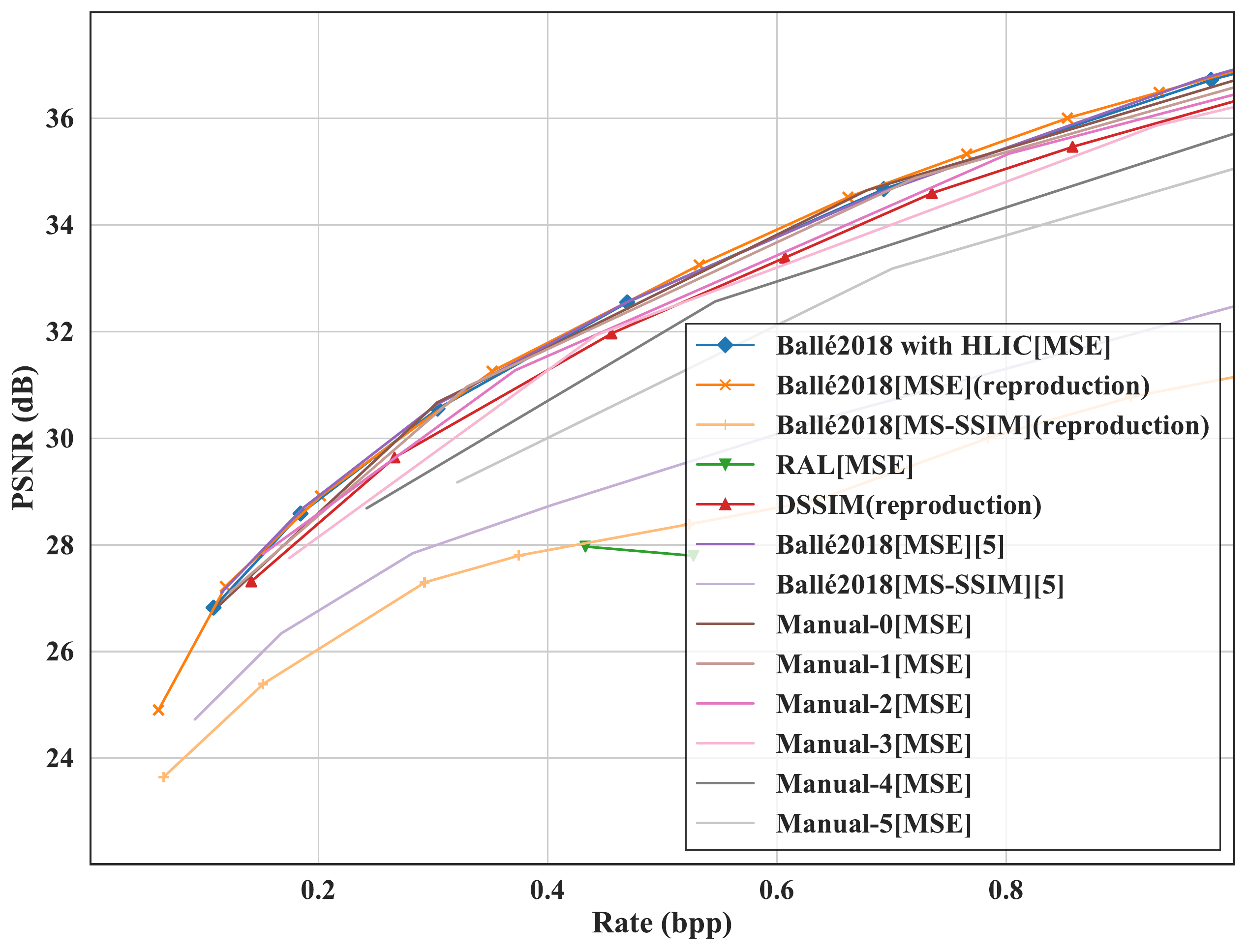}
    \end{minipage}
    \begin{minipage}[t]{0.5\linewidth}
        \centering
        \includegraphics[width=8.5cm]{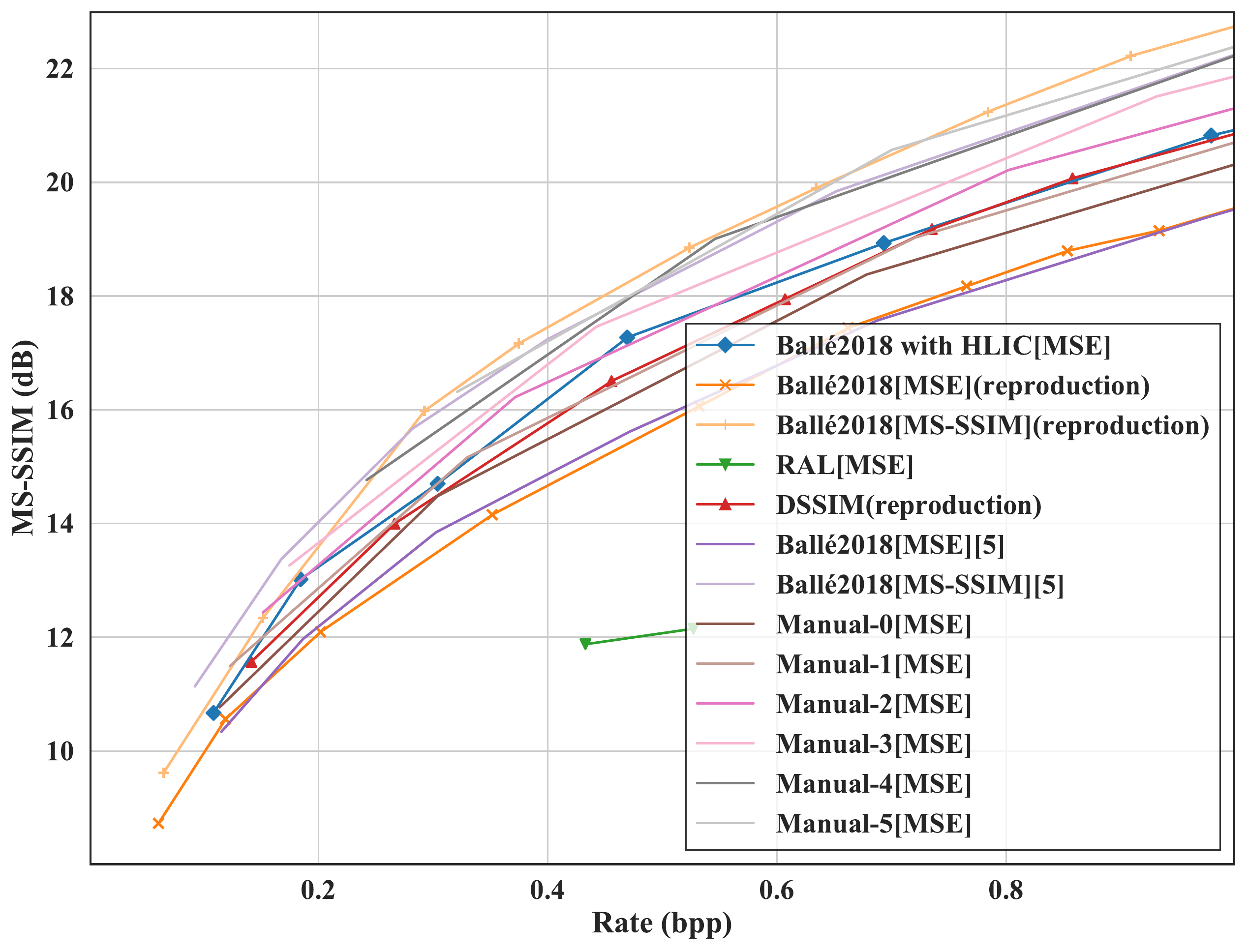}
    \end{minipage}
    \quad
     \begin{minipage}[t]{0.5\linewidth}
        \centering
        \includegraphics[width=8.5cm]{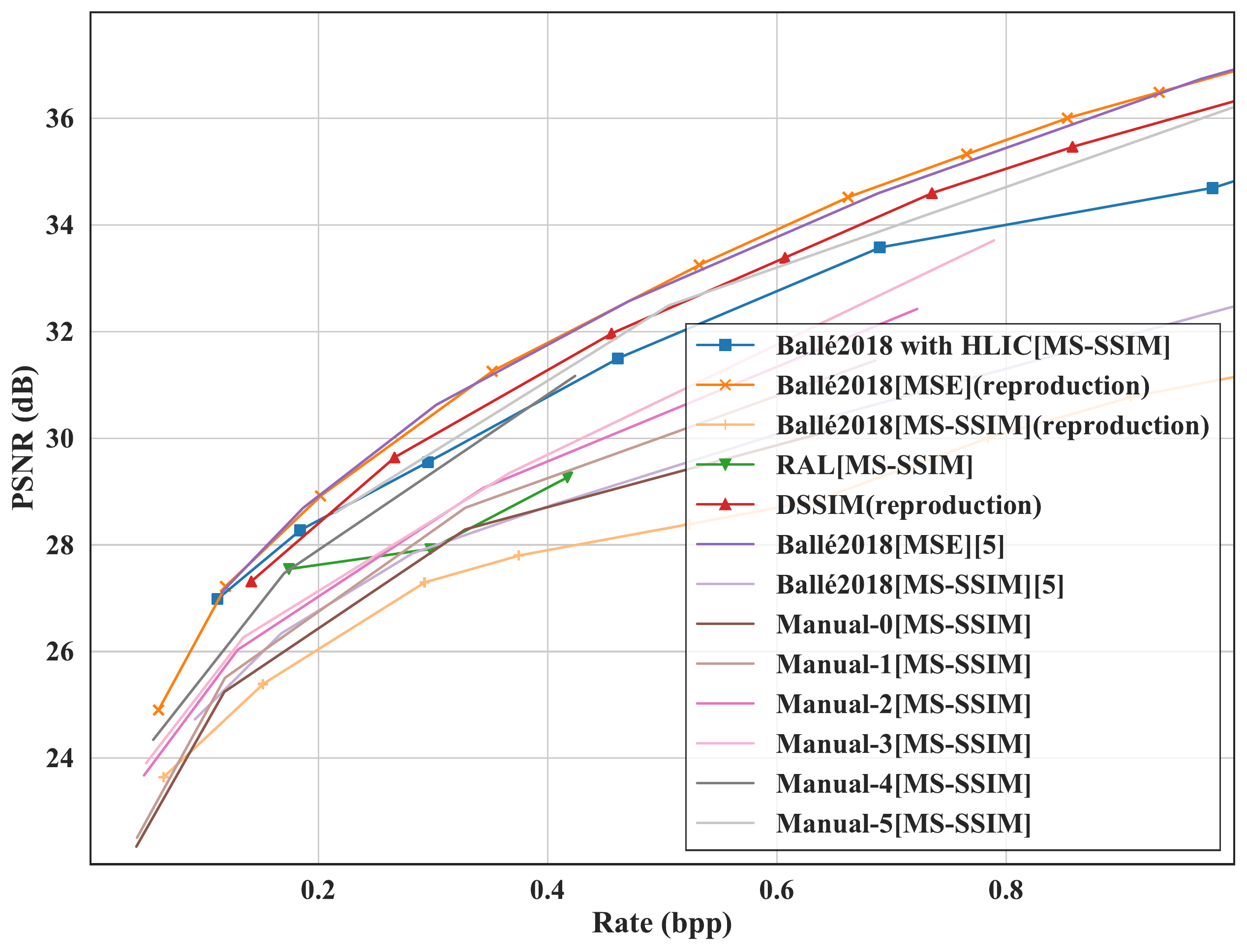}
    \end{minipage}
    \begin{minipage}[t]{0.5\linewidth}
        \centering
        \includegraphics[width=8.5cm]{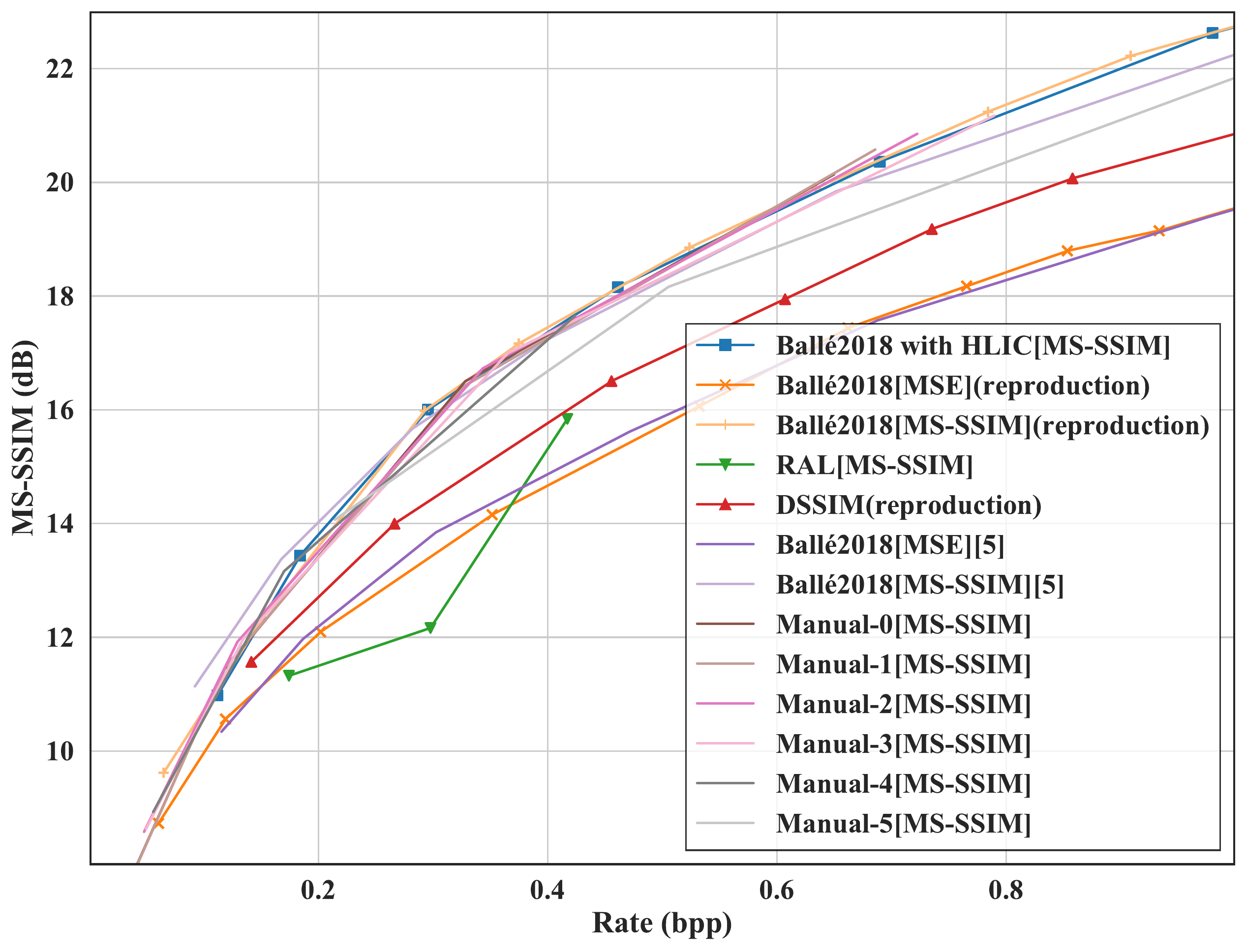}
    \end{minipage}
    \caption{
        RD comparison with several hand-crafted loss functions on Ball\'{e}2018~\cite{balle2018variational}.
    }
    \label{fig:gg18CMH}
\end{figure*}

\begin{figure*}
     \begin{minipage}[t]{0.5\linewidth}
        \centering
        \includegraphics[width=8.5cm]{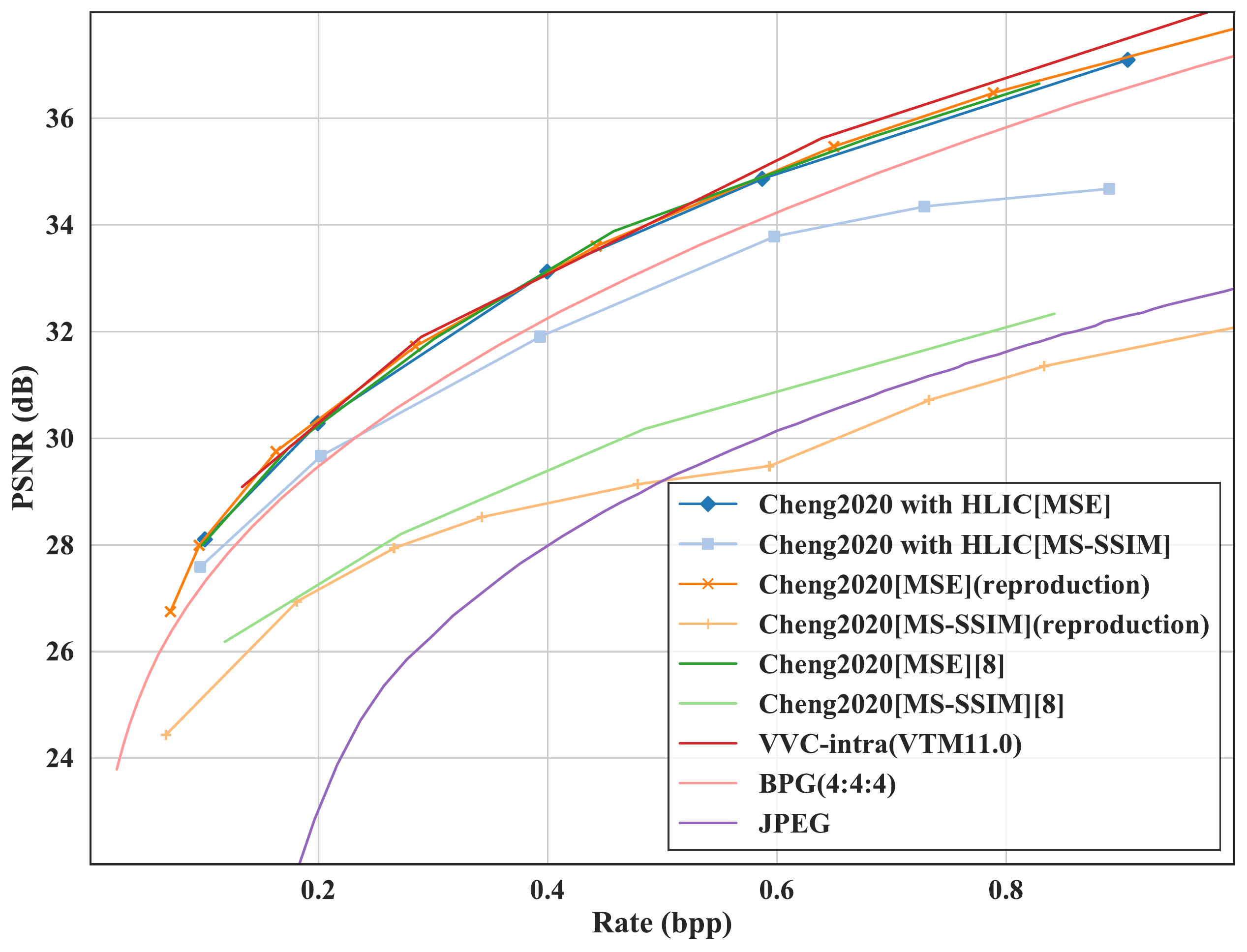}
    \end{minipage}
    \begin{minipage}[t]{0.5\linewidth}
        \centering
        \includegraphics[width=8.5cm]{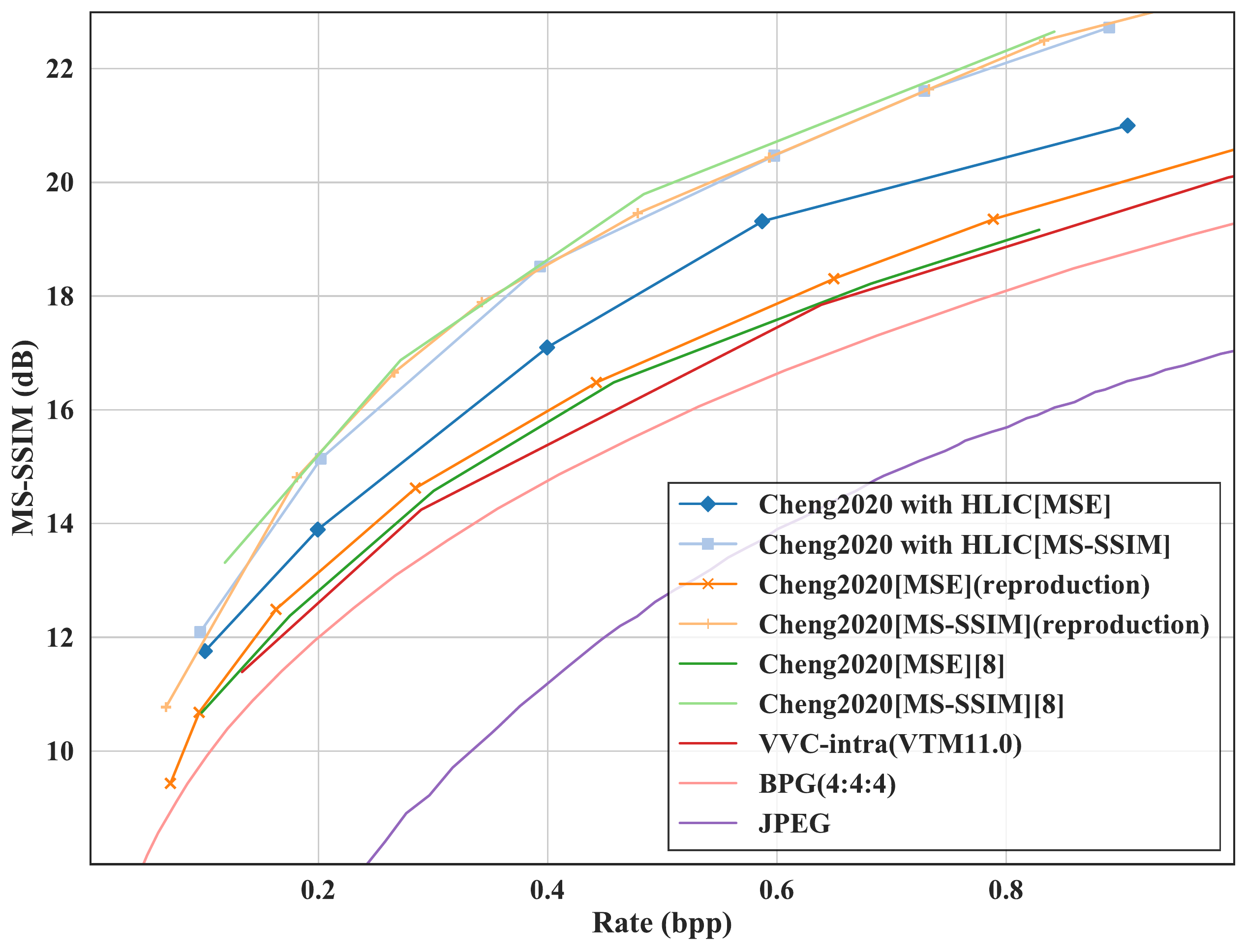}
    \end{minipage}
    \caption{
        Harmonized RD curves evaluated on Cheng2020~\cite{cheng2020learned}.
    }
    \label{fig:cheng20CMH}
\end{figure*}

\label{section: controllable}
We compare our proposed method with the result of hand-crafted hybrid loss similar to~\cite{zhou2018variational, zhou2019end}, which can also harmonize the optimization. For manually optimizing MS-SSIM preferentially while considering MSE (marked as Manual-id[MS-SSIM] in Figure~\ref{fig:gg18CMH}), the hand-crafted distortion weight $\lambda$ is selected as:
\begin{equation}
        \begin{aligned}
            \lambda_{\text{MS-SSIM}} &= \frac{120}{4^{j}} 
            , \lambda_{\text{MSE}} = \frac{0.0128 \times 2^{i}}{4^{j}} \\
            \lambda_{\text{bpp}} &= 1, (i = 0,\dots,5;j=0,\dots,3)
        \end{aligned}
        \label{eq:man_ms}
\end{equation}
 For optimizing MSE preferentially while considering MS-SSIM (marked as Manual-id[MSE] in Figure~\ref{fig:gg18CMH}), the distortion weight $\lambda$ is selected as:
\begin{equation}
        \begin{aligned}
            \lambda_{\text{MS-SSIM}} &= \frac{3 \times 2^{i}}{4^{j}}
            , \lambda_{\text{MSE}} =\frac{0.08}{4^{j}} \\
            \lambda_{\text{bpp}} &= 1, (i = 0,\dots,5;j=0,\dots,3)
        \end{aligned}
        \label{eq:man_mse}
\end{equation}
 where $i$ controls the tradeoff between PSNR and MS-SSIM, $j$ corresponds to different bit rates.
 
The preference in HLIC is controlled according to eq.~\ref{eq:reward2}, which is designed to optimize one metric preferentially while giving attention to the other metric. HLIC[MSE] emphasizes the second term in eq.~\ref{eq:reward}. HLIC[MS-SSIM] emphasizes the third term. RAL denotes using the reward as loss function directly instead of using reinforcement learning, as it is actually differentiable. Compared with hand-crafted harmonized loss function, the preferred balance point can be found by our HLIC more conveniently as we do not have to try different $\lambda$ combinations like that in eq.\ref{eq:man_ms} and eq.\ref{eq:man_mse}. RAL does not perform well and the training is not stable, showing the necessity of applying reinforcement learning. We also compare HLIC with DSSIM designed in~\cite{johnston2018improved} and achieve better performance overall. Similar controllable and harmonized RD performance by our HLIC on Cheng2020~\cite{cheng2020learned}
is shown in Figure~\ref{fig:cheng20CMH}.
\subsection{Qualitative Results}
\begin{figure*}
    \centering
    \includegraphics[width=16cm]{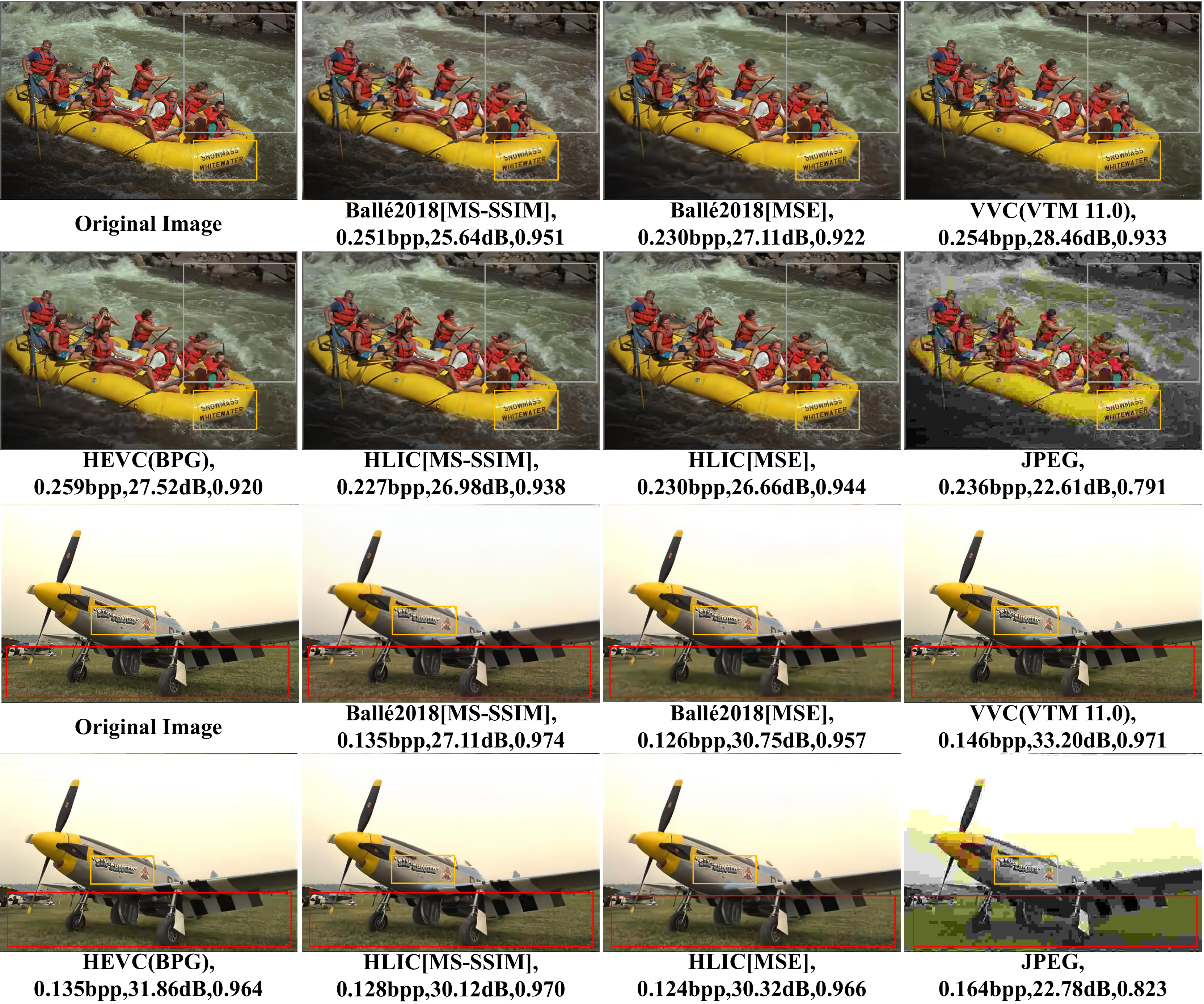}
    \caption{
         Visualization of reconstructed images $kodim14$ and $kodim20$ from Kodak dataset (better zoomed in and viewed in color).
    }
    \label{fig:ablation}
\end{figure*}
\begin{figure}
    \centering
    \includegraphics[width=8.5cm]{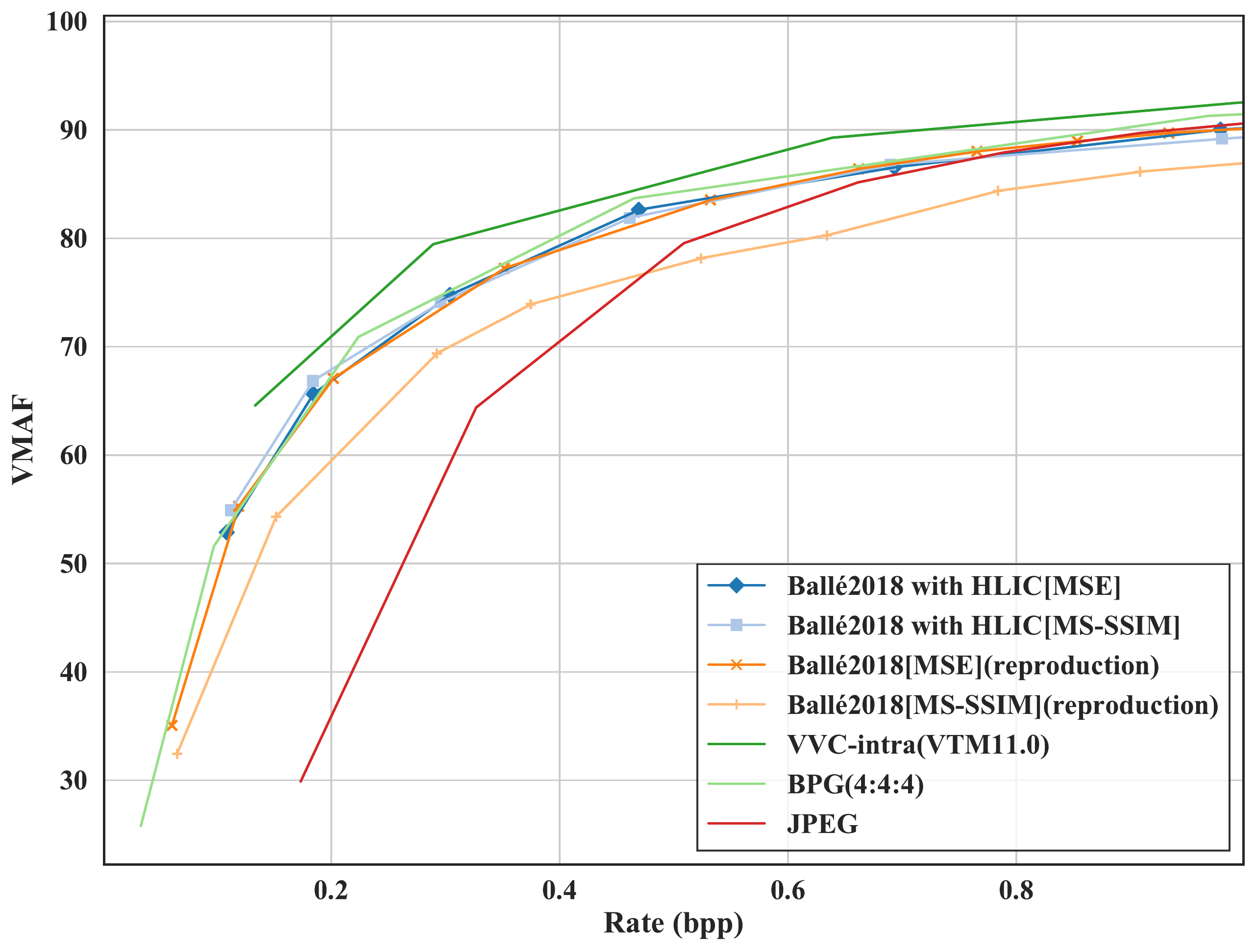}
    \caption{
        VMAF of the reconstructed images from our purposed methods.
    }
    \label{fig:vmaf}
\end{figure}
Our method alleviates image quality problems in baseline models which optimize MSE and MS-SSIM separately. 
We visualize some reconstructed images from the baseline models and proposed HLIC models obtained in section~\ref{section: controllable}.

The first two rows in Figure~\ref{fig:ablation} show the reconstructed images of $kodim14$ in Kodak dataset, with approximately 0.2 bpp and a compression ratio of 120:1. Although the bpp of Ball\'{e}2018[MS-SSIM] is higher and maintains reasonable texture in the water, serious color shift is produced and the text is blurred. Ball\'{e}2018[MSE] significantly reduces textures. Our two HLIC models keep reasonable textures without causing color shift.

The last two rows in Figure~\ref{fig:ablation} show the reconstructed images of $kodim20$, with approximately 0.1 bpp and a compression ratio of 240:1. Ball\'{e}2018[MS-SSIM] leads to artifacts and blur in the text on the plane while Ball\'{e}2018[MSE] overly smooths the grass. Our two HLIC models perform better. 
We also draw VMAF with respect to bpp in Figure~\ref{fig:vmaf}, and our HLIC results significantly improve the VMAF compared with Ball\'{e}2018[MS-SSIM]. 

\section{Discussion}
We propose to Harmonize optimization metrics in Learned Image Compression (HLIC) by an automatic method of online loss function adaptation. Results show the improvement in evaluation metrics as well as visual quality.

Note that although we try to harmonize PSNR and MS-SSIM in learned image compression by proposing HLIC, there still exists metric preference inevitably. As demonstrated in section~\ref{section: controllable}, we cannot achieve best PSNR and best MS-SSIM in one model simultaneously. It is an interesting future direction to investigate more representative optimization metrics and to what extent they can be harmonized by HLIC-like methods with elaborate adaptation space.

While we only demonstrate the effectiveness of harmonizing optimization metrics for learned image compression, it can be extended to various low-level vision tasks where manually designed loss functions often do not correlate well with human perception. 

{\small
\bibliographystyle{ieee_fullname}
\bibliography{main}
}

\end{document}